\documentclass[Afour,sageh,times, doublespace]{sagej}
\usepackage{color} 
\usepackage{moreverb,url}

\usepackage[colorlinks,bookmarksopen,bookmarksnumbered,citecolor=red,urlcolor=red]{hyperref}

\newcommand\BibTeX{{\rmfamily B\kern-.05em \textsc{i\kern-.025em b}\kern-.08em
T\kern-.1667em\lower.7ex\hbox{E}\kern-.125emX}}

\def\volumeyear{2017}

\begin{document}

\runninghead{Kelchtermans and Tuytelaars}

\title{How hard is it to cross the room ? \\
- Training (Recurrent) Neural Networks to steer a UAV}

\author{Klaas Kelchtermans\affilnum{1} and Tinne Tuytelaars\affilnum{1}}

\affiliation{\affilnum{1}KU Leuven, ESAT-PSI, iMinds, Kasteelpark Arenberg 10, bus 2441, B-3001 Leuven, Belgium}


\email{klaas.kelchtermans@esat.kuleuven.be}

\begin{abstract}
This work explores the feasibility of steering a drone with a (recurrent) neural network, based on input from a forward looking camera, in the context of a high-level navigation task. We set up a generic framework for training a network to perform navigation tasks based on imitation learning. It can be applied to both aerial and land vehicles. As a proof of concept we apply it to a UAV (Unmanned Aerial Vehicle) in a simulated environment, learning to cross a room containing a number of obstacles. 
So far only feedforward neural networks (FNNs) have been used to train UAV control. To cope with more complex tasks, we propose the use of recurrent neural networks (RNN) instead and successfully train an LSTM (Long-Short Term Memory) network for controlling UAVs. 
Vision based control is a sequential prediction problem, known for its highly correlated input data. The correlation makes training a network hard, especially an RNN. To overcome this issue, we investigate an alternative sampling method during training, namely window-wise truncated backpropagation through time (WW-TBPTT). 
Further, end-to-end training requires a lot of data which often is not available. Therefore, we compare the performance of retraining only the  Fully Connected (FC) and LSTM control layers with networks which are trained end-to-end. 
Performing the relatively simple task of crossing a room already reveals important guidelines and good practices for training neural control networks. Different visualizations help to explain the behavior learned.
\end{abstract}

\keywords{Navigation control, imitation learning, UAV, drone, deep learning, LSTM, finetuning, sequential prediction, obstacle avoidance, autonomous, indoor flight, simulation}

\maketitle

\section{Introduction}
\label{sec:intro}

The revival of neural networks in the form of deep learning has been at
the basis of significant breakthroughs in various application domains, including speech recognition and computer vision. In the context of robotics, however, adoption of deep learning methods seems to happen at a slower pace and is met with more skepticism. 
Several complicating factors may be at the basis of this phenomenon. 
First and foremost, robotics involves embodied physical systems. This implies 
that datasets cannot so easily be shared, as they tend to be robot-specific. 
Data collection is thus considerably more time consuming, even more so since we are dealing with active systems, which interact with their environment.
This high burden in terms of data collection hampers progress, given the data-hungry nature of deep neural networks. 
Recently, however, \citep{cad2rl} has demonstrated that a control network for single-image obstacle avoidance trained solely in simulation can generalize to the real world. 
In this work, we experiment in a simulated environment, focusing on the basics, assuming that the step to the real world can be solved in a similar manner.

On top of the difficulty of data collection, there are the traditional objections with respect to neural networks, such as the non-convexity of the parameter spaces resulting in local minima; the lack of interpretation of what the network has actually learned; and the large number of hyperparameters which need to be set. 

On the other hand, neural networks hold a lot of promises, also for robotics applications. In particular, they cope well with high-dimensional input data; they can learn the optimal representation for a given task, instead of relying on handcrafted features; and they are universal function approximators. Finally, they are highly non-linear, as is the world and (presumably) the control needed in such world. 

Most importantly, the introduction of (deep) learning in robotics holds the promise of going 
beyond the currently dominating model-driven, metric approach to robotics. 
Indeed, while such model-driven approaches work well for low-level control and/or for robot operations in a highly structured and controlled environment, they reach their limits when it comes to more flexible systems which need to adapt to their environment in a smart way, or need to interact with people. 
For high-level tasks, it may be easier to just show examples of how one would like the robot to behave, and learn directly from such data, rather than handcrafting features, finite-state machines, rules and algorithms implementing the intended behavior. In an ideal setting, learning a new task then boils down to collecting representative data, together with the desired outputs. 

In particular, our long term goal is a framework, in which one can train an unmanned aerial vehicle (UAV) to perform
a wide range of high-level navigation tasks, based on {\em imitation learning}.
That is, the system learns how to perform a task based on training data, in which an expert steers the drone and demonstrates the desired behavior, similar to apprenticeship learning \citep{helicopter}.
Note that
we exclude low-level tasks such as attitude control like \citep{aggressive}, for which we rely on standard algorithms which come with most commercial drones. 
Instead, we focus on the higher-level task of navigation, i.e. steering the drone. High-level tasks we would like our framework to learn could vary from flying a fixed route, avoiding obstacles,  passing through a door or following a corridor to 
tracking a person, recording a high-jump or inspecting a windmill. 

Moreover, we want to achieve this goal using a {\em forward looking camera as the only sensor}. Indeed, experience from human pilots performing such tasks shows that the input from such camera over time contains enough information. Cameras can be made very light, both physically and power consumption wise. They are also not limited to a certain range unlike active sensors.
Additional sensors might simplify some problems, yet bring extra weight which reduces the flight time. 

At test time, the system should then be able to steer the drone and perform the task, based on 
the video input stream only, under conditions similar to those seen at training time. 

\begin{figure}[t]
\begin{center}
\includegraphics[width=0.6\linewidth]{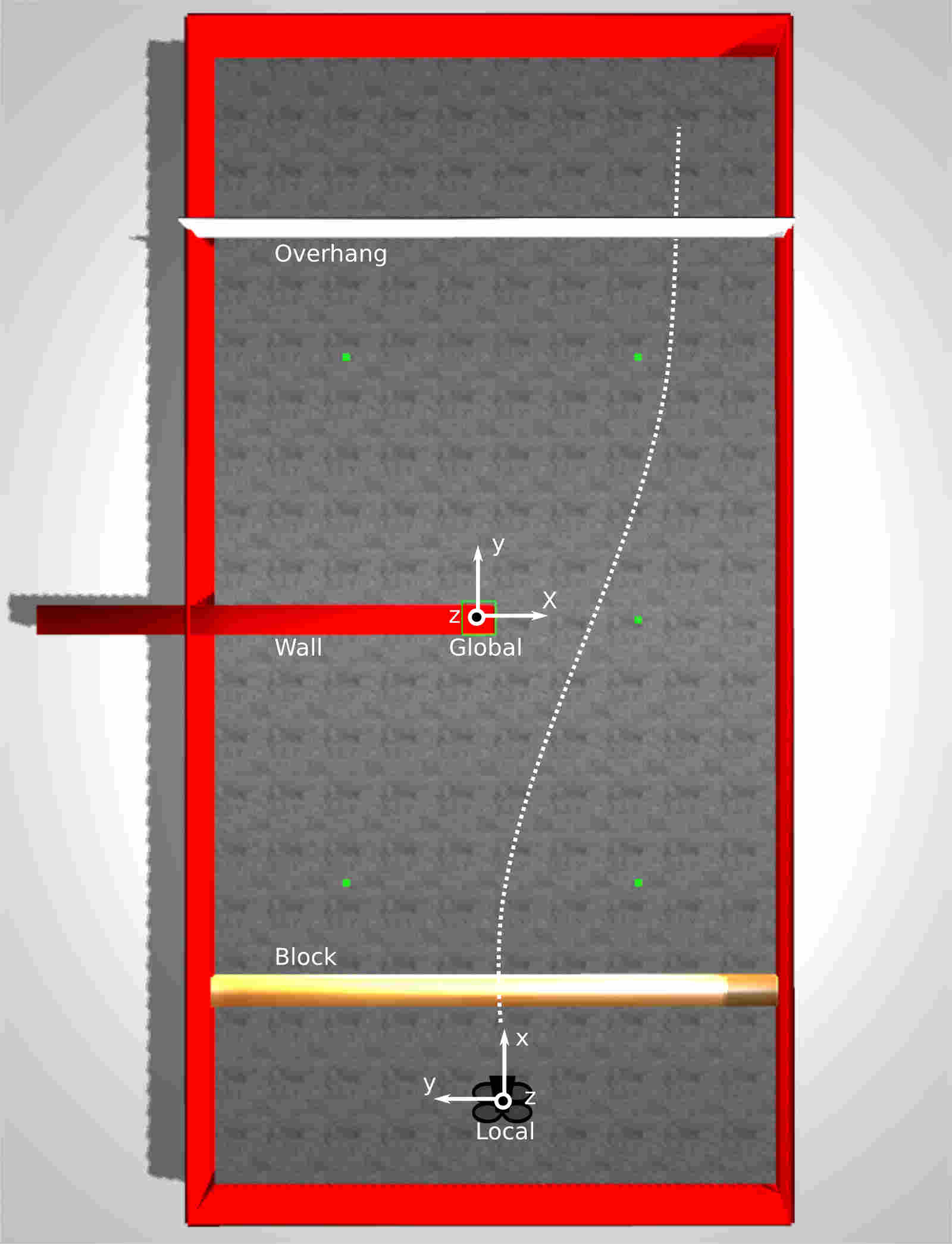}
\end{center}
   \caption{A top view of the Room Crossing challenge in simulation. The global coordinate system is defined in the center of the room, while the local coordinate system is defined in the center of the drone. The dotted line gives a possible path for the drone to follow: over the block, next to the wall and under the overhang.}
\label{fig:theroom}
\end{figure}

For now, as a first step in that direction, we focus on a single, relatively simple task: traversing a room, with three
known obstacles (a bump in the floor, a wall on the left 
or right 
hand side and an obstacle hanging from the ceiling) -- see Figure~\ref{fig:theroom}. The order in which the obstacles appear, is fixed 
for some experiments and variable for others; their dimensions (i.e. the height of the bump/overhang and the length of the wall) always vary. 

This somewhat mimics a setting where a drone flies in an unknown environment, but is given high-level instructions so knows roughly what to expect 
or how to cope with certain obstacles. 
As indicated earlier, to easily generate different rooms and for ease of experimentation, we limit ourselves to a virtual world only. 
Moreover, instead of manually flying the UAV in this world to generate training data, we place additional virtual sensors on the drone, based on which a behavior arbitration algorithm for this particular task can be developed relatively easily. This algorithm serves as expert in our experiments. This saves time during experimentation and ensures reproducibility of the results. 

Within this setting, we then explore the impact of various design choices and the effect of different training methods. In particular, some of the questions we try to answer in this paper include: 
\begin{enumerate}
\item { Is it advantageous to use a network with a memory (RNN instead of CNN) ?}
\item { What is the best strategy to cope with the high correlation between samples in sequential data?}
\item { How to deal with the state space distribution shift when switching from the expert to the student ? }
\item { Is it necessary to train end-to-end, or is retraining the last layer(s) sufficient ? }
\item { What are some guidelines / best practices to ensure quick learning ? }
\end{enumerate}

We focus especially on the first question, i.e. the introduction of networks with a memory. Applications of neural networks for robot control in the real world are mainly limited to memory-free feedforward networks~\citep{dronetrails, MPC, treeavoiding}.
Yet we believe that for high-level tasks, some form of memory or inner state is actually needed. In our setting, one cannot expect the forward-looking camera to always provide enough information to take the proper action without such context.  
The memory provided by an LSTM can help the control network to 
take the right decisions.  
For instance, the network can learn robustness to temporal distortions like delays which are common in real-time applications. Or, it can remember the drone is in the middle of a complex maneuver (e.g. overtaking or moving away from an obstacle), even if the current input is ambiguous. Besides, the state can be extracted from both temporal as well as spatial features and there is no theoretical boundary on the time-span of the memory.

The main difficulty with sequential prediction problems, like navigation control, is the high correlation between the samples. This makes training a network, especially an LSTM, challenging. In this work we 
study how to successfully train an LSTM. In this context, we propose a new sampling scheme, which we coin window-wise truncated backpropagation through time (WW-TBPTT). This addresses the second question.

There is another issue, specific to imitation learning.
In a naive approach, training data is collected offline, with the expert controlling the drone. This data is used to train a model which is then applied at test time. However, navigation control is an active system. Once the student, in our case the neural network, provides the control, it is likely to make mistakes never made by the expert. This brings the drone in situations never seen during training. Special strategies are needed to learn how to recover from these mistakes. We explore different methods to cope with this state-space distribution shift: we experiment with DAgger~\citep{DAgger}, which stands for data aggregation
and we test the use of recovery cameras during training, as used by \cite{pomerleau} and \cite{car}. This addresses the third question.

It has been shown that convolutional neural networks (CNN) are capable of learning to estimate the optical flow \cite{flownet} or depth \cite{depth} from an RGB image. With end-to-end learning the network can define a proper state representation combined with the proper control. Using an RNN allows to build both temporal as well as spatial representations. Yet end-to-end learning is especially data-hungry.
To tackle the fourth question, we compare different networks, either trained end-to-end or starting from a pretrained network and retraining only the last control layers. For the latter, we build on a standard image classification network. 

The fifth and final question about guidelines and best practices is addressed throughout the entire paper and experimental setup. 

The main contributions of this paper can then be summarized as follows:
i) the successful demonstration of UAV control based on LSTM in a navigation task using imitation learning, including a novel sampling method during training; 
ii) a synthetic dataset and baselines for a specific use case, namely learning to cross a virtual room containing various obstacles, with a behavior arbitration algorithm as expert; and
iii) a study of how to train neural networks for such a control task, resulting in guidelines and good practices which may be helpful for other researchers.

The remainder of the paper is organized as follows. First, we describe related work (Section~\ref{sec:related}). Next, we give more details on the standard network architectures and training methods we will be building on (Section~\ref{sec:background}). In Section~\ref{sec:method}, we first give more details on the overall setup (Section~\ref{sec:framework}), 
the particular tasks we are addressing and the dataset used (Section~\ref{sec:data}). Then, we explain the behavior arbitration system we will be using as expert in our experiments (Section~\ref{sec:behaviorarbitration}). After that, we propose an alternative sampling method for LSTMs, window-wise truncated backpropagation through time (WW-TBPTT) (Section~\ref{sec:sampling}). 
Section \ref{sec:impldetails} covers the implementation details.
In Section~\ref{sec:expres} we describe our experimental results, and Section~\ref{sec:conclu} concludes the paper.

\section{RELATED WORK}
\label{sec:related}

\paragraph{Vision-based navigation}
Many systems tackle the navigation problem by simultaneously localizing the vehicle and building a map of the environment (SLAM) solely based on RGB images.
But these systems fail as soon as the tracking fails, e.g. when there are no clear features in the camera view~\citep{LSD}.

In \citep{monocularflight} a control for autonomous navigation in the forest is implemented based on a forward and a downward looking camera, a sonar sensor and IMU data. The images are sent to a base station on which depth is estimated. From the depth, a 3D reconstruction is made and used for motion planning. This is computationally very expensive and therefore unfeasible to run on board.

In other work~\citep{opticflow}, the control is based on the difference in optic flow over a wide-view camera.

In this work we train a neural network to incorporate these different complex tasks. There is no need for an explicit 3D reconstruction as the network inherently learns to use the 3D information obtained from the image to navigate the drone correctly. This is computationally much less expensive and can happen on board. Moreover, there is no need to explicitly choose the type of information provided to the control. 

\paragraph{Control systems based on Neural Networks}
Training a control network with solely RGB images as input has already been demonstrated in 1990, by \cite{pomerleau}. 
In that work, an FNN was trained online from a set of shifted and rotated images. This important work showed that networks are capable of performing a restrictive task like following a road. Also, it showed the need for recovery data in the training set. The network contained only 5 hidden units and 30 discrete output units. The computational power of today allows us to work with more complex networks and continuous control.

It is much more difficult to pilot an aerial vehicle than it is to keep a car on the road, given the same amount of congestion. \cite{dronetrails} trained a deep CNN to follow forest trails. A big dataset of trails recorded from 3 cameras was created. One camera facing forward was annotated with the control of going straight. Two cameras pointing sideways had annotated control to compensate for the different orientations. The deep network was able to classify the images with high accuracy.

In this work we train a network to apply continuous control. This means that we change the machine learning problem from a classification task to a regression task. 

\paragraph{Control systems based on imitation learning}
In \citep{treeavoiding} the control is learned by imitation learning. 
They use SVM (Support Vector Machines) with as input a combination of image features, optic flow, IMU data and the previous applied control. After an initial offline learning stage, the control is applied in an online fashion under supervision of a human expert. If the control is going to crash the human supervisor takes over. The SVM is then retrained on the aggregated dataset. This principle is called DAgger which stands for {\em data aggregation}~\citep{DAgger}. Manually annotating and supervising the controller during training can be tedious and costly. In this work we propose an automated manner to overcome this difficulty. 

Another difficulty encountered by~\cite{treeavoiding} was that once the obstacle is out of the field of view of the drone, the navigation control stops avoiding this obstacle while it might still be in flying range. This was often the reason for a crash. In this work we train both FNNs as well as RNNs. RNNs have a memory which can be especially useful in these situations.

In~\citep{MPC} a supervised guided policy search is applied to drones with the aid of Model Predictive Control which uses extra sensor input in order to fully observe the current state of the drone. An FNN is trained to follow a corridor and to avoid thin obstacles based on the input from laser range sensors.
Our work differs in the sense that our supervisor is defined with behavior arbitration \citep{behaviorarbitration} and we use RGB input only. 
Adding memory could avoid the need for extra sensors.

The tasks in the previous examples were relatively primitive. The image itself contains the necessary information to make the right decision. In this work we look at a higher level of control which comes closer to trajectory following.

\cite{car} show how a CNN can learn to behave in a wide variety of situations. They train a very deep network of 9 layers for which a large amount of real world data, 72 hours of driving, was obtained. With the aid of a simulator the data was further augmented. The simulator interprets the real data and creates a model of the perceived environment. By shifting and interpolating between sideways looking cameras, a variety of orientations and driving behavior is obtained and annotated with corresponding control labels. This driving behavior differs from the expert's behavior, providing examples which are unlike the perfect expert's behavior.

Neural networks need a lot of annotated data. \cite{synthia} have made a virtual environment from which a very big annotated dataset was obtained. They train a control network to drive a car autonomously on a dataset containing less than 8 percent of real data. In the same spirit, we see it fit to explore the training behavior of different control networks first in a simulated environment. Once the control behaves properly the step to the real world only needs a relatively small amount of extra training data from the real world.

In contrast to previous work, we do not restrict the control problem to 1 dimensional steering. Our task also contains obstacle avoidance in the vertical plane, applying 2D control signals. Avoiding objects in the vertical direction can be very effective.

All the networks mentioned above are trained end-to-end. This means that both the feature extraction and the control behavior are learned simultaneously. It also means that the data required to avoid overfitting increases. 

In~\citep{focus} pretrained CNN models for object proposals are used to find free space in order to avoid obstacles. This is one way to overcome the big data demand, although it comes close to a handcrafting solution.
In the area of image recognition it is a common practice to use off-the-shelf CNN features \citep{off-the-shelf} for representing the high dimensional input image. In this work we explore if this common practice can be applied to learning a control and reducing the big data demand.

\paragraph{Control systems based on reinforcement learning}
In the full reinforcement learning problem an agent with no prior knowledge of the task, needs to find the right policy which maximizes the reward. This is a very devious way of learning a desired policy or behavior. In contrast, in imitation learning an expert demonstrates the desired behavior in order to set the learning in the right direction.

A big breakthrough in training control networks for full reinforcement learning, was the work of \cite{atari}. They succeed in training a deep CNN to play several Atari games at super human performance with deep Q-learning. At the input they stack 4 consecutive frames of the player's view. The control output of the network is directly fed to the game. An important difference with our work is that the screen of an Atari game is much better in representing the current state of the agent and shows better what the agent needs to do next, compared to the first person view coming from a forward looking camera on a drone spawned in a room in which it should follow a certain trajectory.

Finally, limited work has been done by \cite{bakker} in training an LSTM for controlling an agent in a reinforcement learning toy example. This work focuses on online learning, i.e. with the training data provided sequentially. This makes the training procedure very slow for big networks. 

\begin{figure*}[t]
\begin{center}
\includegraphics[width=0.8\linewidth]{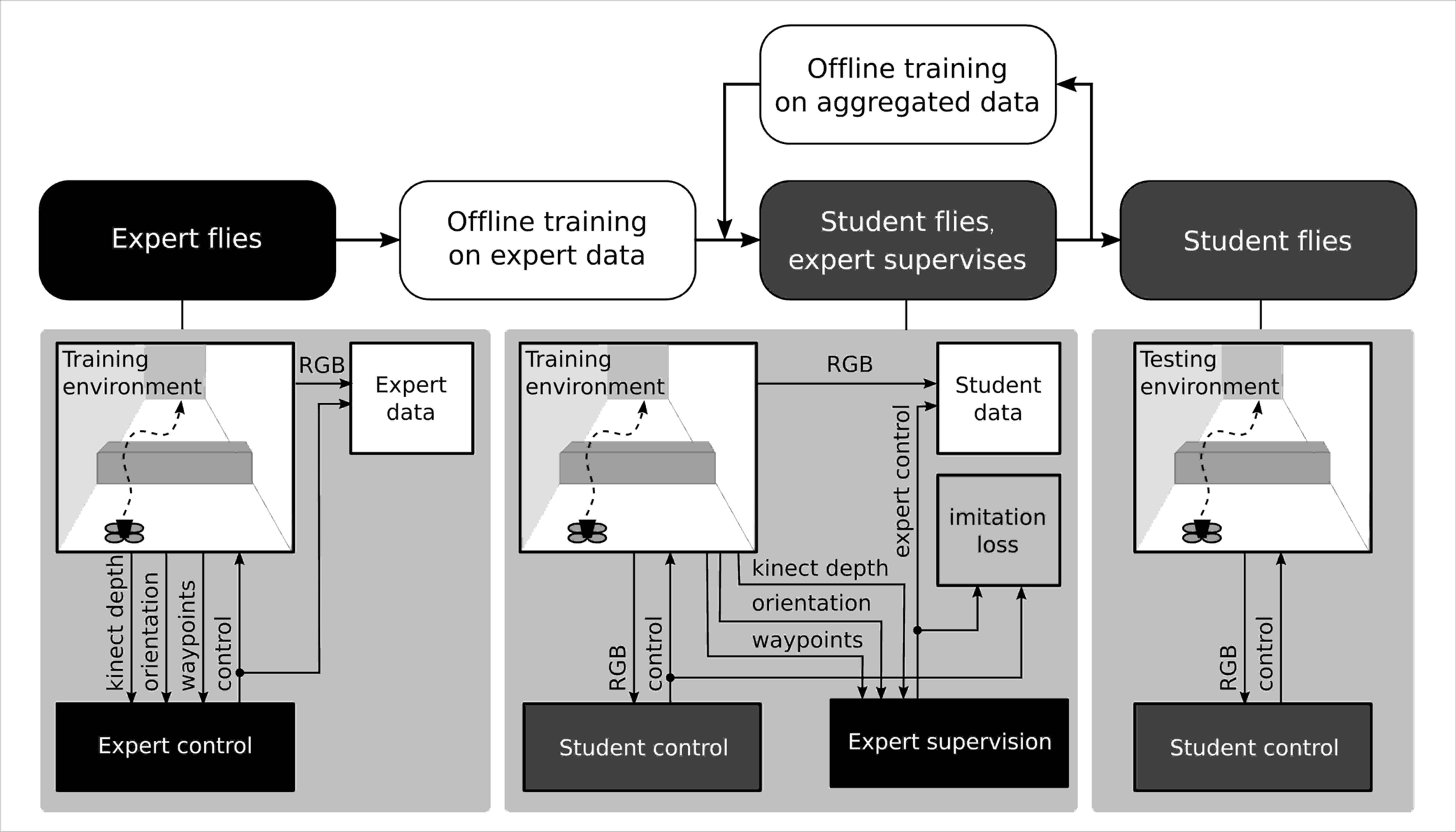}
\end{center}
   \caption{The generic framework for tackling imitation learning for neural networks, applied to our Room challenge. An automated expert, in this case a behavior arbitration scheme, uses the input of extra sensors to perform the task and generate expert data for offline training (left). The automated expert also supervises the behavior of the student, in this case a neural network. The student is trained and online tested iteratively (middle). The data during testing is aggregated to the dataset according to the DAgger principle \citep{DAgger}. In the final test setup the student performs the task without supervision of the expert (right).}
\label{fig:imitationmethod}
\end{figure*}

\section{Background}
\label{sec:background}

\begin{table}
\small\sf\centering
\caption{Overview of common abbreviations. \label{Table Abbreviations}}
\begin{tabular}{lll}
\toprule
UAV & Unmanned Aerial Vehicle\\
NN & Neural Network\\
FNN & Feedforward Neural Network\\
CNN & Convolutional Neural Network\\
RNN & Recurrent Neural Network\\
MLP & Multilayer Perceptron\\
FC & Fully Connected\\
LSTM & Long Short-Term Memory\\
BPTT & Backpropagation Through Time\\
F-BPTT & Fully Unrolled Backpropagation Through Time\\
TBPTT & Truncated Backpropagation Through Time\\
WW-TBPTT & Window-wise Truncated BPTT\\
S-TBPTT & Sliding Truncated BPTT\\
RTRL & Real-Time Recurrent Learning\\
CEC & Constant Error Carousel\\
SLAM & Simultaneous Localization And Mapping\\
\bottomrule
\end{tabular}\\[11pt]
\end{table}
This section introduces concepts of deep learning like CNNs, LSTMs and truncated BPTT (backpropagation through time). In table \ref{Table Abbreviations} an overview is given of the used abbreviations. If the reader is familiar with these concepts, feel free to skip this background section. The information is based on the tutorial by \cite{silver}.

Deep learning originates from neural networks with multiple layers, also called {\em multilayer perceptrons}. They are {\em fully connected} in the sense that each node of a layer gets weighted input from the output of all nodes in the previous layer. This can lead to a very big amount of weights, especially when the input is high dimensional like an image.

The weights are learned using {\em backpropagation}. In a forward pass through the network, the input flows to the output. The output is compared with the desired output or label using an appropriate loss function. The gradients of this loss function towards the weights in the network are calculated with the chain rule in a backward pass. The optimization of the weights by applying gradients is called {\em gradient descent}. There are many different optimizers for applying gradient descent. In this work, we use the ADAM optimizer \citep{ADAM}.

The number of weights in a network can be reduced by sharing them. Weight sharing can be done in the spatial domain with a so called {\em convolution layer}. This is a stack of convolution filters which in general are much smaller than the input, for example $3 \times 3$ or $5 \times 5$. The output of this layer are the activations of each convolution filter. The convolution operation is often followed with a non-linear activation function like with multilayer perceptrons. The result is a directed graph, also called a {\em convolutional neural network} (CNN). They are very common in computer vision and have shown to be very powerful in complex computer vision tasks like face or object recognition.

A second way to share weights is in time over an input sequence. In order to interpret sequences of input the network needs to have a memory which is possible with {\em recurrent neural networks} (RNN). In an RNN the nodes are connected over different time steps. By unrolling the network over these time steps one gets a directed graph just like a normal feedforward neural network. Gradient descent can be applied to this unrolled directed graph in order to find the appropriate weights. This is called {\em backpropagation through time} (BPTT)~\cite{truncated bptt}.

In order to train the memory of the RNN over multiple time steps, the network needs to be further unrolled. In general the gradients tend to explode or to vanish when backpropagated over many time steps. This makes training long term memory hard. \cite{LSTM} introduced the LSTM model (Long Short-Term Memory) to overcome this issue. They work with a constant error carousel (CEC) which keeps the error gradient constant over different time steps. It works with an inner cell containing the state of the node and different gates controlling the influence of the input and the influence on the output. The functionality is very well explained by \cite{LSTM blog}.

Unrolling the network over the full input sequence ({\em full BPTT}) often results in stability issues during training. The network is therefore often only unrolled over the time steps of a subsequence of the input. This subsequence can be seen as a time window which slides over the data starting at the beginning of the data sequence where the stored values of the recurrent network are all zero. In the next step, this time window is shifted one step in time over the input sequence and again applied to the unrolled network. The stored values of the unrolled network in this step is not the initial state but the state after the first time step. The sliding truncated BPTT method (S-TBPTT) is very well explained by \cite{truncated bptt}.

\section{Method}
\label{sec:method}
In this section we first describe the generic imitation learning framework (Section~\ref{sec:framework}). Then, we zoom in on the Room Crossing task on which the performance of the different control networks are compared (Section~\ref{sec:data}). Next, the automated expert based on behavior arbitration is explained (Section~\ref{sec:behaviorarbitration}).
Finally, we give the details of the window-wise truncated backpropagation through time (WW-TBPTT) by comparing this method with the normal sliding truncated BPTT (Section~\ref{sec:sampling}).

\subsection{Imitation Learning framework}
\label{sec:framework}

The generic framework for imitation learning we are using is illustrated in Figure~\ref{fig:imitationmethod}. It already includes the basic idea of data aggregation, in the style of DAgger~\citep{DAgger}.

In imitation learning the control is learned from demonstration by an expert. In reinforcement learning this is also referred to as {\em guided policy search}~\citep{visuomotor}.
The student, in our case the neural network, has to learn to mimic the behavior of the expert. In the rest of this section, we use the term 'student' to refer to the control network.

In practice the expert is often a human pilot or operator.
When a lot of training data is required, as is the case for training a deep neural network, this is very costly.
Moreover, when using DAgger iterations, we have online experiments with a human in the loop which can become very time consuming. Finally, such experiments are hard to reproduce. Therefore, we propose to use an {\em automated expert} instead. The key insight here is that by adding extra sensors the complexity of the navigation task can be reduced significantly - to the point where it becomes relatively straightforward to implement a control algorithm which solves the task. 
In a real-world setting, this could be a motion capture system in a training arena for drones.
In a simulated environment as many sensors (e.g. depth \& pose) can be used as required for building a control algorithm which can perform the task automatically.

The actual imitation learning then consists of three stages. In a first stage, training data is collected in an offline fashion, by giving the control to the automated expert and have him perform the task a number of times. The sensor input available for the student (in this case the RGB camera) is recorded and labeled with the control applied by the expert.  
This is shown in the left block of figure \ref{fig:imitationmethod}.

Based on this primary expert data, an initial model is trained offline in a supervised manner. At this point, the student has only learned to copy the behavior of the expert. However, since the expert does not make any mistakes, the student has not learned how to react to or anticipate on mistakes. In reinforcement learning, this is referred to as a {\em state space shift}. The state space of the expert differs from the state space of the student, yet the student has only learned how to behave in the state space of the expert~\citep{DAgger}.
Therefore it is necessary to let the student perform the task under the supervision of the expert. This happens in the second stage of our framework.

In this stage, the control is given to the student. The student tries to perform the task using only the subset of the sensor input available at test time (e.g., the RGB camera). This is shown in the center block of figure \ref{fig:imitationmethod}.
The expert annotates this new set of data with controls which it would have applied while using all the sensor inputs (e.g., depth \& pose).
In \citep{treeavoiding}, the supervision is done manually which is very time consuming and prone to errors.
Deep learning has already proven to work for big annotated datasets though in robotics this can be a very time consuming task. With this setting based on an automated expert, we overcome the need for human annotation, making the framework very powerful.

After the first flight of the student the fresh data is aggregated to the primary dataset according to the DAgger algorithm~\citep{DAgger}.
The new dataset contains mistakes which the student has just made and is most likely to make again if it wasn't retrained. 
This makes the data extra relevant for the student. This is somewhat akin to hard negative mining.
Compare it, for example, with a student who learns to drive a car. The first step is an offline demonstration. Later the student tries to drive under supervision of the expert. The student learns from his mistakes based on the corrections made by the expert.

Offline retraining of the student and online performance evaluation and supervision are then iterated a number of times until the student's behavior is close enough to the expert's behavior. Notice that in this setup the student will never get better than the expert. 
If this is required, we refer to the full reinforcement learning problem where the student continues training, thereby maximizing the reward obtained with the task. 
Even then, it is recommended to start the full reinforcement learning with a control network which acts similar to an expert in order to start the search to the optimal policy in the right direction.

Finally, when the student's control is sufficiently similar to the expert's control, we can move to the third stage, where the student can fly the drone autonomously.

While this is our generic setup, we show in the experimental section \ref{sec:expres} that the DAgger technique is not as robust as expected.

\subsection{Navigating across the room}
\label{sec:data}
The task of the navigation control in our study is to fly a UAV across a room with known obstacles. This challenge might seem trivial at a first impression though it incorporates different behaviors. The room is made in the Gazebo simulation environment~\citep{gazebo}. 

The drone is spawned at one side of a long room. Once it has taken off and reached the proper height, the navigation control steers the drone to the other side of the room. In the room there are 3 obstacles: a block, a wall and an overhang. These objects come in different sizes, so the control algorithm has to adapt to the sensor input. The block and the overhang require vertical maneuvers while the wall requires a horizontal maneuver.

The vertical maneuvers of the block and the overhang can be seen as reactive behavior, in the sense that it can react last-minute. At the moment the drone observes it is close to an obstacle, it simply needs to translate in local z-direction. The horizontal maneuver of the wall, on the other hand, is more tricky.
In our setting the drone should not translate sideways. This ensures that the drone is always flying in the direction of the forward looking camera. So in order to avoid the wall it needs to turn around the local z-axis resulting in a yaw angle. From a control point of view this is much harder.

In order to avoid the wall the navigation control should steer the drone from the beginning towards the opening next to the wall. This is a behavior that comes closer to path planning than the low-level reactive obstacle avoidance. 

The expert crosses the room with three controls varying between $-1$ and $1$: translation in local x and z-direction and rotation in yaw. The other three command values are annotated with zero. More details about our Room Crossing dataset are given in Section \ref{sec:room}.

\subsection{Behavior Arbitration as Automated Expert}
\label{sec:behaviorarbitration}

In our setting we use a powerful control technique, called behavior arbitration, to define the automated expert \citep{behaviorarbitration}.
Behavior arbitration allows the combination of different behaviors using different sensors to be combined in a natural and logical manner.
In our setting, the behavior arbitration scheme combines two behaviors. 
The first behavior is reactive obstacle avoidance based on depth images obtained from a Kinect. 
The second behavior is a higher level path planning defined by waypoints. The current orientation of the drone is given by the simulated environment. This behavior turns the drone in the direction of the next waypoint. 
This implementation comes from the idea that mobile robots are often provided with a GPS signal which can give the orientation to the next point but fails to avoid the obstacles. 
Our task is intentionally kept simple.
The framework with behavior arbitration allows more complex behaviors to be added to the scheme like door-crossing, window-crossing, dynamic object avoidance, ... as shown in \citep{behaviorarbitration}.

\subsection{Time window sampling, an uncorrelated training method for LSTMs}
\label{sec:sampling}

\begin{figure*}[t]
\begin{center}
\includegraphics[width=0.6\linewidth]{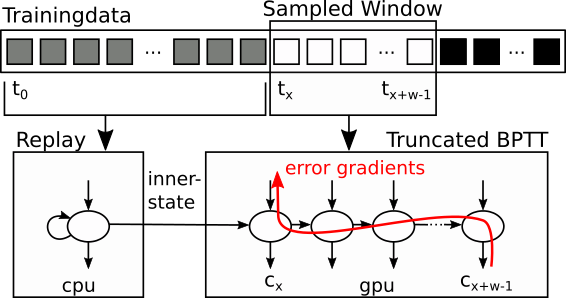}
\end{center}
   \caption{Window sampled of size w in training sequence.}
\label{fig:window}
\end{figure*}

From a reinforcement learning point of view, the goal of a sequential prediction problem like navigation control is to predict the best control given the current observed state of the agent. 
The state is in this case the input image \citep{atari}.
In the case of a feedforward CNN, the decision is made based on the current image only. 
In the case of a 3D CNN, a fixed number of consecutive images are concatenated as done by \cite{atari}.
Recurrent Neural Networks are different in the sense that the decision is based on both the current input as well as the memory contained in an inner cell state. This stored value depends on previous input. 

As shown by \citet{atari}, it is necessary to decorrelate the input samples by randomizing the order. 
For training LSTMs we build further on this idea by sampling time windows of varying length at random locations in the data.

LSTMs (Long-Short Term Memory) are specifically designed to be able to learn longer temporal correlations by using an input-, output- and forget-gate as explained in Section~\ref{sec:background}. These gates ensure that the backpropagated error does not explode or disappear as is a known problem for regular RNNs. 
There are different ways to train an RNN/LSTM. The two most popular methods are Real Time Recurrent Learning (RTRL) \citep{RTRL} and Backpropagation Through Time (BPTT). BPTT is by far the most used due to the high computational cost of RTRL~\citep{truncated bptt}.

With BPTT, the recurrent network is unrolled over time resulting in a feedforward neural network. The unrolled network is shown in Figure \ref{fig:window}. There are three ways of using BPTT.
A first approach is feeding the image sequence to the network in its full length (F-BPTT). This allows the error gradients to flow from the last control output back until the first input frame. It requires the RNN to be fully unrolled over the different timesteps which is memory-wise very demanding and trains very slowly. 
A second approach, sliding truncated BPTT (S-TBPTT), is to let a time window of fixed length (eg. 20 steps) slide over the image sequence starting at time $t_0=0$ when the network has the initial zero-state. The stored value after the first output is kept for the next sliding window position. The advantage of this approach is that the network only needs to unroll over 20 steps. The disadvantage is that the data is always fed in the same order, making the input correlated and introducing a sequential bias. 
We propose a third method called window-wise truncated backpropagation through time (WW-TBPTT) in which the windows are sampled at random positions in the data with different random window lengths. This method is explained in Figure \ref{fig:window}. 
Picking the windows at random positions in the data sequence decorrelates the samples. This results in higher variance in the training data which leads to slower convergence during training though it overcomes the sequential bias of S-TBPTT. 

As explained in Figure \ref{fig:window} a time window is picked at position $t_x$ and has length $w$. In order to apply truncated BPTT the stored value of the RNN at time $t_{x-1}$ needs to be known. Therefore the network needs to replay the input sequence up until time $t_{x-1}$. The stored value at that time step is then fed together with the time window to the unrolled network. Each epoch the positions of the different time windows are picked randomly. The different stored values can be obtained with several threads on a CPU as the network does not need to unroll.

The recurrent network is unrolled over $w$ steps. This allows the error gradients to float back over a maximum time span of $w$ steps. The network is, in other words, restricted to find temporal correlations over maximum $w$ steps. The maximum memory span of the RNN will be $w$ steps if it is trained with truncated BPTT. 

Different tasks might require different memory spans. Following a certain trajectory in a maze of corridors might require the full trajectory length. In other words this can be an important limitation of RNNs trained with truncated BPTT.

If the GPU is not capable of unrolling the network, a simple workaround is to downsample the input sequences. In our experiment the images are obtained at a rate of 10 frames per second. RNNs are expected to be able to generalize over time.

Even if the network can generalize over different frame rates, the window size should be chosen carefully. As the time window size increases or when the data is downsampled, the number of unoverlapping time windows in the training data decreases. This can result in severe overfitting due to too few different samples. From experience we know that a model with around 100 000 parameters is best trained with at least 1000 unoverlapping samples. In case of a fully unrolled BPTT, the expert trainingset needs to contain at least 1000 demonstrations. This is clearly not feasible for a human expert. With our method, we can sample 50 windows of 20 time steps from a normal sequence of 1000 frames, requiring only 20 demonstrations by an expert. This reveals a tradeoff between the maximum possible memory span and the chance of overfitting due to lack of data.

\section{Implementation Details}
\label{sec:impldetails}
In most of our experiments the control network is trained by taking a feature extracting network and only training the last two FC or LSTM layers. The reason is that training only the last decision layers goes much faster than training the full network end-to-end. We also implement an end-to-end convolutional network for comparison. The specifications of the Room Crossing dataset made by the automated expert are explained at the end of this section.

\subsection{Feature extracting network}
\label{sec:featextract}
Due to the high dimensionality ($640$x$480$x$3$), it is unfeasible to feed a raw image directly to an LSTM or FC. We extract features from the input image by taking the activations of the 3rd pooling layer of the pretrained Inception v3 network \citep{Inception}. This is a convolutional network trained on the Imagenet large scale image classification challenge \citep{imagenet}. The features are known to be a generic and compact representation of the image \citep{off-the-shelf}. The dimensionality is reduced to an array of $2048$.

\subsection{Control networks}
\label{sec:controlnets}
We compare the performance of three control networks: FC, 5-FC and LSTM.  The Fully Connected (FC) is a multilayer perceptron with 2 layers of 400 hidden units. The 5-FC takes the concatenation of the features of 5 consecutive frames as input. It has 2 layers of 100 hidden units in order to keep the complexity similar. The LSTM has 2 layers of 100 hidden units resulting again in a similar complexity. 
All networks have one extra fully connected layer with 6 units, one for each control output (3 translations and 3 rotations). In our Room Crossing dataset only 3 control outputs are non-zero as explained in Section~\ref{sec:data}. The control networks are trained with the Adam optimizer \citep{ADAM}. The loss function is the RMS error of the predicted control output and the desired control output averaged over the number of time steps and the training batch. The networks are both implemented in Tensorflow \citep{tensorflow}.

\subsection{End-to-end networks}
\label{sec:endtoend}
\begin{figure*}[t]
\begin{center}
\includegraphics[width=\linewidth]{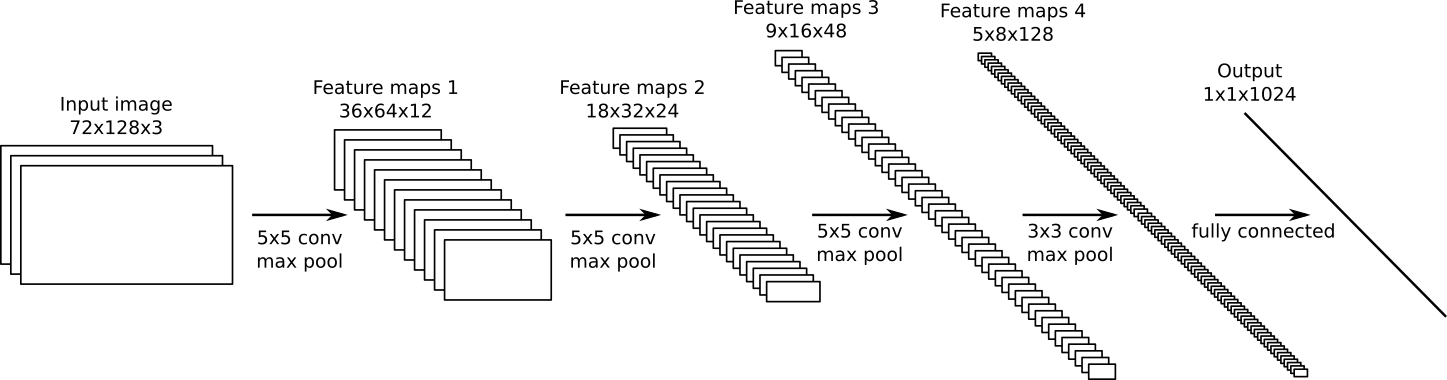}
\end{center}
   \caption{The architecture of the convolutional network. After the output layer follows the control networks described in \ref{sec:controlnets}.}
\label{fig:convnet}
\end{figure*}

The end-to-end networks consist of 3 convolution layers with each a ReLU activation function, followed by a control network like the ones explained above. The networks are trained end-to-end. This means that the errors at the output can flow back up until the first convolution layer. The images are fed to the network at a lower resolution (128x72). The details of the 3 convolution layers are shown in figure \ref{fig:convnet}.

\subsection{The Room Crossing datasets}
\label{sec:room}
We create two versions of our Room Crossing dataset: the \textsl{Room Crossing One} dataset is relatively easy while the \textsl{Room Crossing Two} dataset is more challenging. 
Both datasets consist of RGB images captured from a drone flying through a simulated environment. The environment is built with ROS \citep{ROS} and Gazebo \citep{gazebo}. We use the drone model provided by TUM \citep{TUM}. We added a Kinect depth sensor to be used by the expert as explained in Section \ref{sec:behaviorarbitration}. The drone spawns on one side of the room ($y_{global}=-16$) and needs to get to the other side ($y_{global}=+16$), as shown in figure \ref{fig:theroom}. The room is 40 units in length (global y-direction), 20 units wide (global x-direction) and 4 units high (global z-direction). The expert flies at a constant speed of more or less 0.4 units per second. 

In the \textsl{Room Crossing One} dataset there are 3 consecutive obstacles: a block, a wall and an overhang. By varying the height of the block and the overhang or the cross position of the wall we have created a set of 18 different rooms, ordered from room 0 till 17 with increasing level of difficulty. In particular, the block varies in height between 0.75 and 1.5 units. The wall shifts in global x-direction leaving an opening of 5 to 10 units wide. The overhang has a height of 1.25 units or 1.5 units. 
Training data is collected by having the expert steer the drone across these room, starting from 2 different heights ($z_{global} = 1$ and $z_{global}=2$) and 3 different x-positions (-0.5, 0, 0.5). This results in $3\times 2 \times 18=108$ different training trajectories. Each trajectory is more or less 800 frames long. The orientation of the yaw angle varies a little at random to introduce potential noise that is likely to be there during test time as well.
The online performance is tested on the same 18 rooms with the drone spawned in the middle of the starting points of the training trajectories ($x_{global}=0$ and $z_{global}=1.5$).  

For the \textsl{Room Crossing Two} dataset we increase the challenge on all fronts. First, we limit the initial training set to 5 rooms only, through which the expert flies just once. The starting position is picked randomly in the range of $x_{global}\in[-0.5:0.5]$ and $z_{global}\in[1:2]$. For 3 supplementary DAgger iterations we have added 3 times 2 new rooms. In order to increase the challenge further, the online performance is tested not just on these existing rooms, but also on 4 completely new rooms, unknown to the student (i.e., never seen during training). All 15 rooms again have a block, an overhang and a wall of varying size as obstacles. However, the order of the obstacles varies, as well as the lighting and the texture on the walls and obstacles. The performance of the networks is tested in both known rooms crossed by the expert while making the initial training data and unknown rooms, always using a fixed starting point ($x_{global}=0$ and $z_{global}=1.5$).

The \textsl{Room Crossing Two} dataset is intentionally kept small in order to represent better a real world imitation learning scenario, even if this may lead to overfitting.

After publication, the datasets will be provided to be used as a benchmark for training navigation control.

\begin{figure}[t]
\centering
  \includegraphics[width=2.1cm]{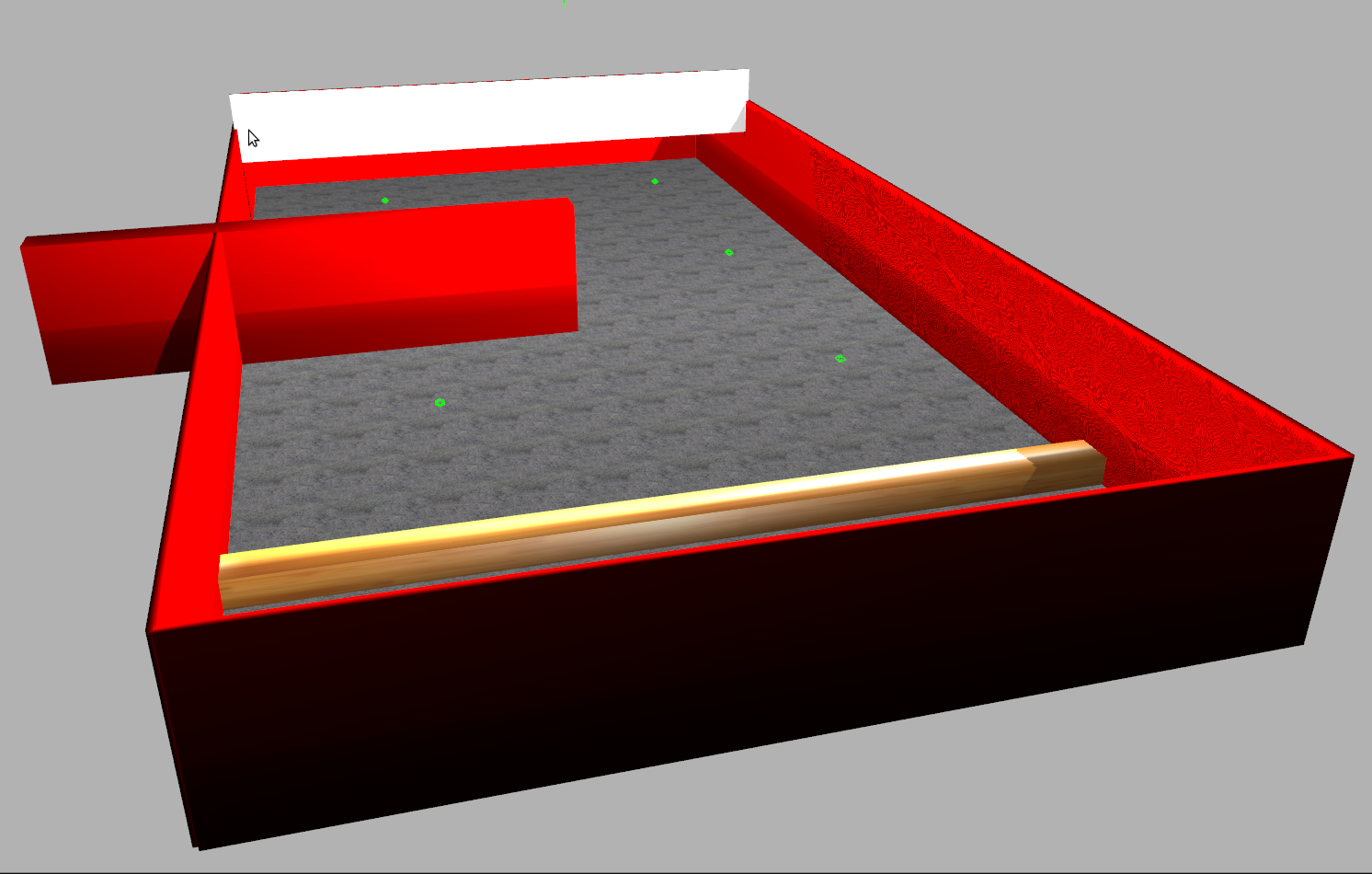}%
\qquad
  \includegraphics[width=2.1cm]{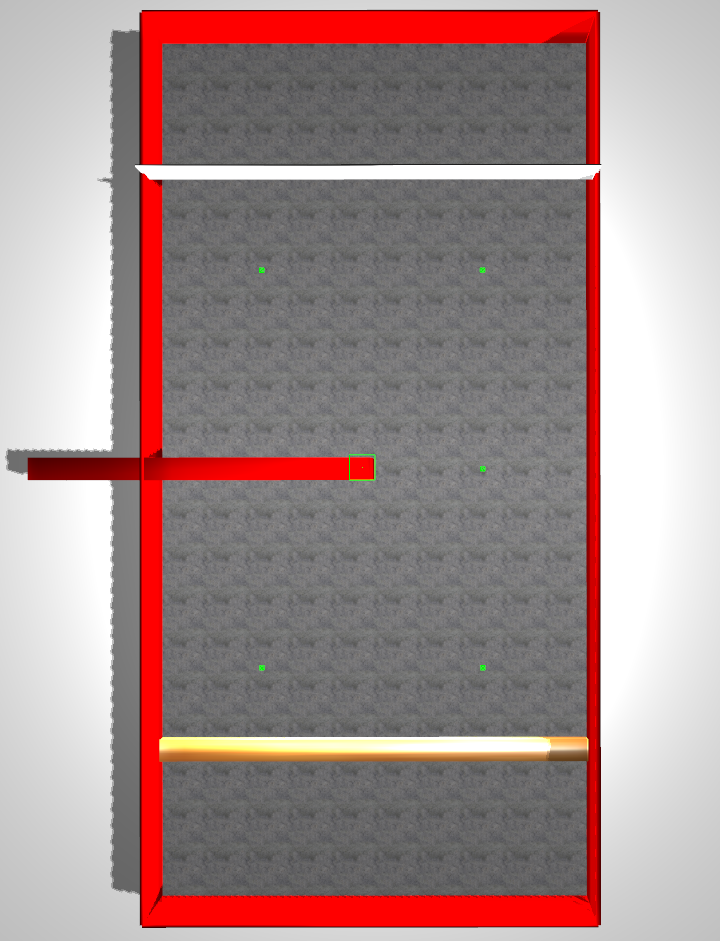}%
\qquad
  \includegraphics[width=2.1cm]{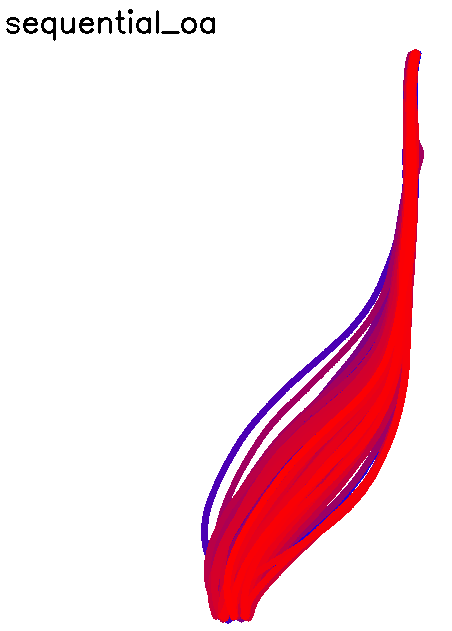}%
\caption{Left and middle: An example room from the Room Crossing One dataset with 3 consecutive obstacles: a block, a wall and an overhang. Right: a tube of different trajectories flown by the expert starting off at different z and x positions.}
\label{fig:room}
\end{figure}
\section{Experimental Results}
\label{sec:expres}
This section gives an overview of the different experiments we conducted in order to answer the 5 questions stated in the introduction. In the first section, \ref{sec:incvsendtoend}, we explore the need of end-to-end training in order to answer question 3 from the introduction. Section \ref{sec:depthvsinc} compares general Inception features with depth features. At the same time this analysis led to some guidelines on how to evaluate different performances giving good practices in answer to question 5. Section \ref{sec:difftrainmethods} explores the different training methods for LSTM answering in this way question 2. Then the most important experiment is addressed in Section \ref{sec:fcvslstm}; namely the use of memory for navigation is explored, revealing the answer to question 1. The last two sections, \ref{sec:recovery} and \ref{sec:dagger}, focus on question 4. They show to what extent the state space shift from expert to student in an imitation learning setting can be tackled with the use of recovery data (\ref{sec:recovery}) or DAgger iterations (\ref{sec:dagger}).

\paragraph{Evaluation criteria} The performance of the networks is expressed with different evaluation measures -- see table \ref{T1}. The task of the control is to fly to the other side of the room. Therefore an obvious evaluation measure is the number of times the control succeeds in doing so. This is expressed as the {\em success rate} in the first column. The second column of table \ref{T1} shows the {\em average imitation loss}. This evaluation measure expresses how close the behavior of the control network is to the expert's behavior. During online performance of the student, or control network, the 2-norm difference between the applied control and the supervised control is kept at each frame as a running average over the last 100 frames. The table shows the average loss over the different rooms. The third column of the table shows the {\em average maximum y-position}. The drone spawns at -16 and needs to get to +16 in global y-direction -- see figure \ref{fig:theroom}. This reveals a more continuous evaluation measure than the success rate. {\em Instead of focusing on a single number, inspection of multiple evaluation measures leads to more insight and easier interpretation of any trends}. This can be noted as a first good practice.

\subsection{Retraining Control Layers vs End-to-end Training}
\label{sec:incvsendtoend}
\begin{figure}[t]
\begin{center}
\includegraphics[width=2.5cm]{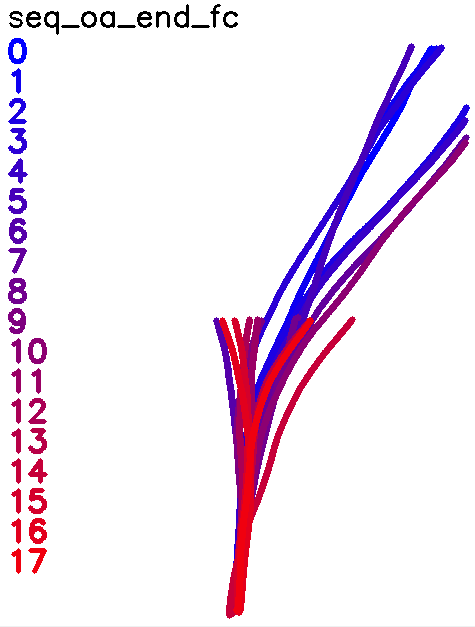}
\quad
\includegraphics[width=2.5cm]{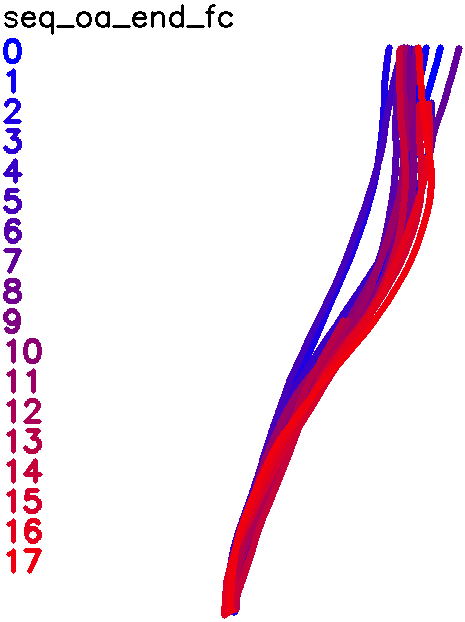}
\quad
\includegraphics[width=2.5cm]{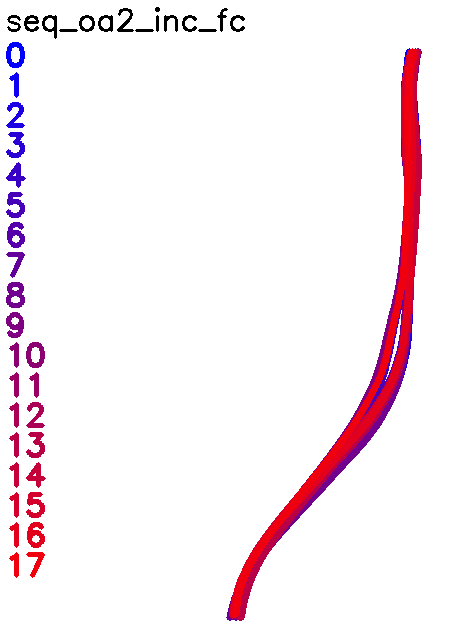}
\end{center}
\caption{The different trajectories steered by a fully connected network trained end-to-end on the Room Crossing One dataset (left), trained end-to-end on an augmented dataset (middle) or trained only the last control layers using pretrained Inception features as input (right).}
\label{fig:incvsendtoend}
\end{figure}

Training control networks end-to-end from scratch can yield optimal representations for the task at hand yet requires a lot more data in order to avoid severe overfitting. Using pretrained convolutional networks for extracting features, on the other hand, avoids this issue. The task on which the convolutional network is trained is very different (classification of objects in real images) than the task for which it is used (indoor navigation of a drone in a simulated environment). Nevertheless the FC control network on Inception features performs much better on the Room Crossing One dataset than the end-to-end FC network, as visible in figure \ref{fig:incvsendtoend}. Augmenting the data for the end-to-end network by a factor 8 (middle of figure \ref{fig:incvsendtoend}) improves the situation somewhat although it is still clearly inferior to the network trained on Inception features.

As shown in table \ref{T1} the end-to-end trained networks succeed only 4 and 12 out of 18 times in reaching the other side of the room, while the FC control with Inception features succeeds always. Besides the severe overfitting, it takes around 40 times longer to train the network end-to-end (6.5h) than only retraining the last layers (10min) on a 2G GPU in Tensorflow \citep{tensorflow} on the small dataset.  For the augmented data, it took us 4 days to train the network  on a 2G GPU. The information extracted by the Inception network pretrained on the Imagenet dataset \citep{imagenet}, contains information which is generic enough to perform the control task. This is an important result as it overcomes the data hungriness and the long training times of end-to-end networks. For the rest of the experiments we never use end-to-end learning; we only retrain the final control networks as explained in \ref{sec:controlnets}.

A third option worth exploring which might give best of both worlds is fine-tuning the Inception network end-to-end after initializing the weights with the pretrained network. This is left as future work.

\underline{Conclusion:} Starting from a pretrained CNN and only retraining the last FC layers seems a good alternative for end-to-end learning from scratch, as it saves training time and is less prone to overfitting so less data-hungry. This holds in spite of the large discrepancy between the task of the pretrained model and the new control task. 


\begin{table*}[h]
\small\sf\centering
\caption{Overview of the performance of networks trained on The Room Crossing One dataset expressed in different evaluation measures. \label{T1}}
\begin{tabular}{llrrr}
\toprule
&Network&Success rate&Imitation loss&Max y position\\
\midrule
Sec \ref{sec:incvsendtoend} &\texttt{end FC} &4/18&1.2& 5.6\\
&\texttt{end FC augmented} &12/18&0.5& 13.5\\
&\texttt{inc FC} &18/18&0.02&16.1\\
\midrule
Sec \ref{sec:depthvsinc} &\texttt{depth FC} &2/18&1.5& 1.1\\
&\texttt{depth LSTM} &3/18&1.4& 2.0\\
\midrule
Sec \ref{sec:difftrainmethods} &\texttt{inc LSTM F-TBPTT} &0/18&1.0& -14.7\\
&\texttt{inc LSTM WW-TBPTT} &18/18&0.05& 16.1\\
&\texttt{inc LSTM S-TBPTT} &18/18&0.02&16.1\\
&\texttt{inc LSTM window-size 5} &18/18&0.1& 16.1\\
&\texttt{inc LSTM window-size 10} &17/18&0.1& 15.9\\
&\texttt{inc LSTM window-size 20} &18/18&0.05&16.1\\
&\texttt{inc LSTM window-size 40} &18/18&0.05&16.1\\
\bottomrule
\end{tabular}\\[11pt]
\end{table*}

\subsection{Clean Depth vs Inception Features}
\label{sec:depthvsinc}
One might expect that feeding the general Inception features trained for Imagenet classification \citep{imagenet} is not as informative as depth. The expert uses clean depth images for avoiding obstacles. So it makes sense to test how well the network performs on down scaled clean depth images from a simulated Kinect.
Somewhat surprisingly, the performance is much worse, as shown in table \ref{T1}. 

\begin{figure}[t]
\begin{center}
  \includegraphics[width=3cm]{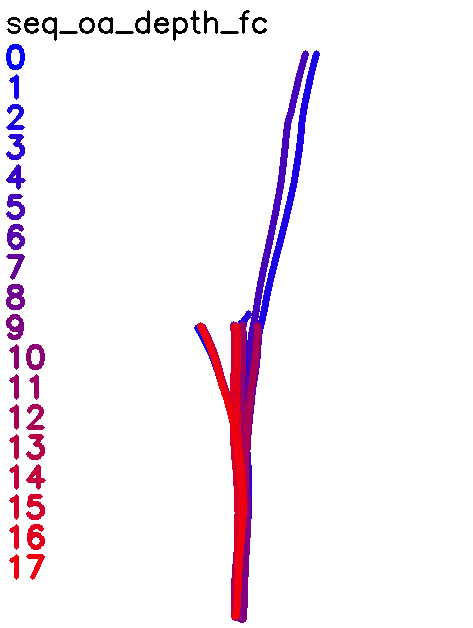}%
\qquad
  \includegraphics[width=3cm]{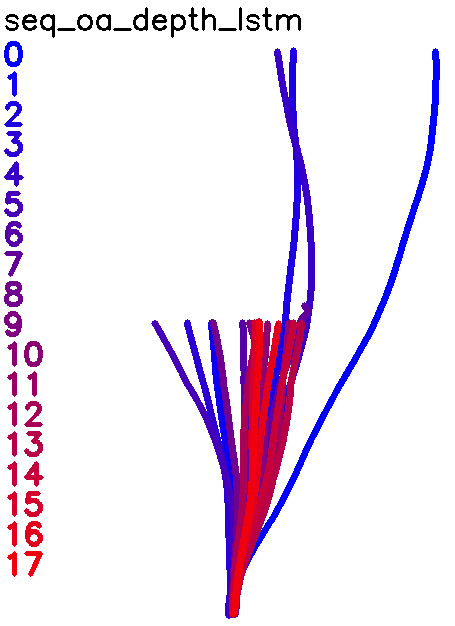}%
\end{center}{center}
\caption{The trajectories steered by FC (left) and LSTM (right) control with clean downscaled depth images as input.}
\label{fig:depth}%
\end{figure}

In order to understand the main reason why the depth input performs worse than the generic Inception features, it is useful to visualize the exact trajectories. Figure \ref{fig:depth} shows the trajectories navigated by the FC and LSTM control networks through the 18 rooms of the \textsl{Room Crossing One} dataset. From these visualizations, it is clear the trained models have difficulty getting passed the wall. 

\begin{figure}[t]
\begin{center}
\includegraphics[width=3cm]{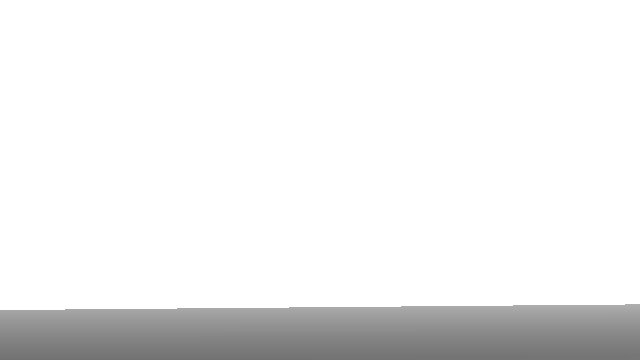}
\quad
\includegraphics[width=3cm]{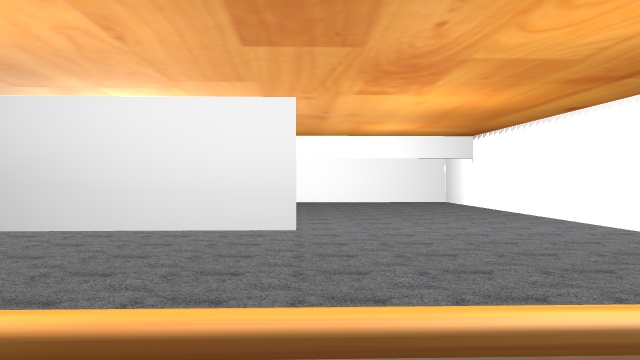}
\end{center}
\caption{Clean depth image (left) does not contain as much information as the RGB equivalent (right).}
\label{fig:depthcomparison}%
\end{figure}

The main reason for this is that the depth images have a limited range (4m), so these models can only react to obstacles when they are close enough -- see figure \ref{fig:depthcomparison}. 
Besides the depth image, the expert also uses the goal direction to the next waypoint, which is in this case the opening next to the wall. 
The Inception features contain perspective information which can be used by the control network to orientate and to provide an initial direction. 
This also shows that the task is harder than reactive low level obstacle avoidance. The drone is expected to anticipate on the wall from the moment it sees it rather than on the moment it is right in front of it. In other words, the control should plan the path. This can also explain why the depth with LSTM performs on all evaluation measures in table \ref{T1} better than the depth with FC control: FC makes the control decision based solely on the current frame so is probably better suited for reactive control than the LSTM.

\underline{Conclusion:} While less related, the generic Inception features seem better suited in this setting than some handcrafted features one might come up with, such as depth. 

\underline{Guideline:}
It is good practice to visualize the trajectories in order to understand what exactly the control network has learned. Likewise, multiple performance measures give a more complete view on the problem leading to better understanding.

\subsection{Different training methods for LSTMs}
\label{sec:difftrainmethods}
\begin{figure}[t]
\begin{center}
\includegraphics[width=2.5cm]{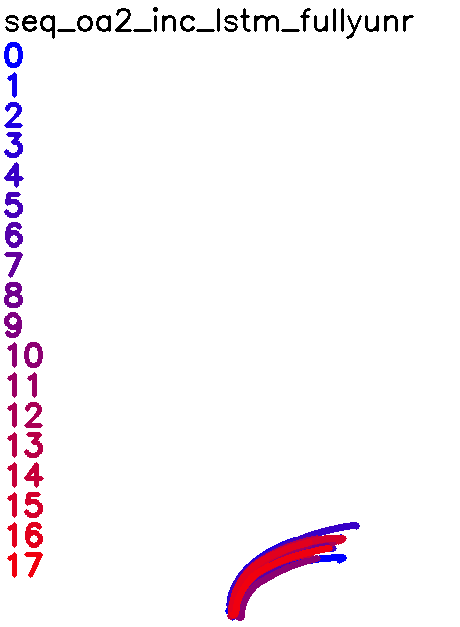}
\quad
\includegraphics[width=2.5cm]{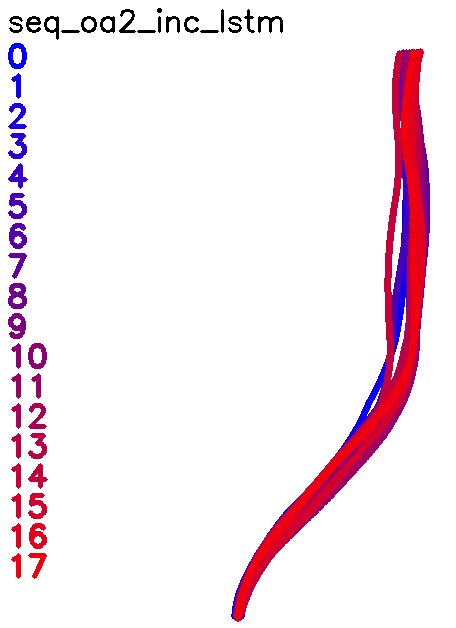}
\quad
\includegraphics[width=2.5cm]{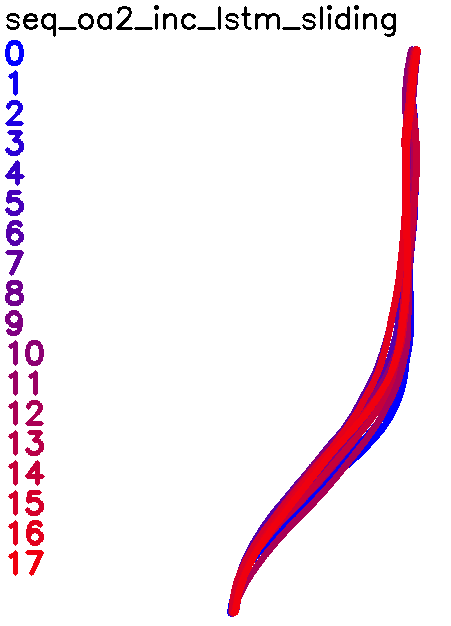}
\end{center}
\caption{The trajectories in the Room Crossing One dataset rooms flown by an LSTM control network trained with 3 different algorithms: F-BPTT (left), WW-TBPTT (middle) and S-TBPTT (right).}
\label{fig:trainmeths}
\end{figure}

Performing navigation based on images is a sequential prediction problem which is hard to train due to the highly correlated data especially for training RNNs. In this section we compare LSTMs trained with different training algorithms. We compare the performance on the \textsl{Room Crossing One} dataset of an LSTM trained with fully unrolled BPTT (F-BPTT), with sliding truncated BPTT (S-TBPTT) and window-wise truncated BPTT (WW-TBPTT) as explained in section \ref{sec:method}. Figure \ref{fig:trainmeths} shows the different trajectories while table \ref{T1} shows the different performance measures. Both the S-TBPTT and the WW-TBPTT method train correctly on the \textsl{Room Crossing One} dataset. The fully unrolled BPTT failed to train properly. This is probably due to the stability issue explained in section \ref{sec:sampling}. Feeding a full trajectory as 1 data sample results in only 100 data samples which is clearly too few for this complex task. Feeding only a truncated sequence allows for a higher variance, necessary to train correctly. Table \ref{T1} shows how the LSTM trained with S-TBPTT has a lower imitation loss as the LSTM trained with WW-TBPTT.


As the window size is an important hyperparameter we explored its influence on the imitation loss in table \ref{T1}. A window size of 20 frames resulted in the lowest imitation loss while it fits still easily on a 2G GPU. In the next experiments we will always use a window size of 20 frames.


Both the control networks trained with S-TBPTT and WW-TBPTT performed perfectly on the \textsl{Room Crossing One} dataset. In the rest of the experimental section, we will work on the more difficult \textsl{Room Crossing Two} dataset, explained in section \ref{sec:room}. The performance is tested online on the set of rooms available in the training data (known rooms) and a set of rooms that the network has not seen before (unknown rooms). The values between brackets in table \ref{T2} correspond to the performance on the unknown rooms. The second and the third row of table \ref{T2} show how the performance of the LSTM trained with S-TBPTT is slightly worse than the LSTM trained on WW-TBPTT, especially on the known rooms.  

With S-TBPTT the training sequences are fed to the network in the same order introducing a bias which is advantageous in the \textsl{Room Crossing One} dataset (where the order of obstacles is fixed) but disadvantageous on the \textsl{Room Crossing Two} dataset (where this is no longer the case). On the other hand the S-TBPTT method converges faster because there is less variance in the data samples. The S-TBPTT  method also takes less training time (25min vs 35min for the Room Crossing Two dataset) because there is no need to calculate the stored value for each training sequence. 

\underline{Conclusion:} The difference between the two sampling methods, S-TBPTT and WW-TBPTT, is relatively small. If the sequential bias of S-TBPTT is harmful for the task, one can best use WW-TBPTT for feeding the training sequences, though this will make the training procedure as well as the convergence slower. In tasks where the control varies at test time in the same order as at training time, the sequential bias can be advantageous in which case the S-TBPTT is the recommended training method.

\begin{table*}[h]
\small\sf\centering
\caption{Overview of the performance of the control networks on the known (unknown) rooms of the Room Crossing Two dataset. \label{T2}}
\begin{tabular}{llrrr}
\toprule
&Network&Success rate&Imitation loss&Max y position\\
\midrule
Sec \ref{sec:fcvslstm}&\texttt{FC} &1/5 (0/4)&3.7 (3.2)&-0.8 (-3.9)\\
&\texttt{5-FC} &1/5 (0/4)&3.8 (2.7)&-1.6 (-3.7)\\
&\texttt{S-TBPTT LSTM} &2/5 (0/4)&1.6 (2.0)& 6.9 (-9.8)\\
&\texttt{WW-TBPTT LSTM} &2/5 (0/4)&1.2 (1.2)& 5.4 (-4.2)\\
\midrule
Sec \ref{sec:recovery}&\texttt{FC with recovery} &3/5 (1/4)&1.4 (1.6)&6.0 (-1.8)\\
&\texttt{5-FC with recovery} &2/5 (0/4)&1.1 (1.9)&5.2 (0.0)\\
&\texttt{S-TBPTT LSTM with recovery} &2/5 (2/4)&0.9 (1.0)& 5.1 (10.6)\\
&\texttt{WW-TBPTT LSTM with recovery} &2/5 (1/4)&0.6 (1.3)& 12.6 (1.2)\\
\midrule
Sec \ref{sec:dagger}&\texttt{LSTM DAgger 1} &2/5 (1/4)&0.6 (1.3)&12.6 (1.2)\\
&\texttt{LSTM DAgger 2} &4/5 (0/4)&0.4 (1.7)&9.9 (-5.0)\\
&\texttt{LSTM DAgger 2 with finetuning} &4/5 (1/4)&1.3 (1.3)&9.9 (2.7)\\
&\texttt{LSTM DAgger 3} &2/5 (1/4)&2.5 (1.6)&3.4 (-2.0)\\
&\texttt{LSTM DAgger 4} &3/5 (1/4)&2.1 (1.7)&9.3 (3.1)\\
\midrule
Sec \ref{sec:dagger}&\texttt{FC DAgger 1} &3/5 (1/4)&1.4 (1.6)&6.0 (-1.8)\\
&\texttt{FC DAgger 2} &3/5 (0/4)&2.7 (2.4)&6.2 (-4.7)\\
&\texttt{FC DAgger 2 with finetuning} &2/5 (0/4)&2.6 (1.6)&3.3 (-1.0)\\
&\texttt{FC DAgger 3} &1/5 (0/4)&2.5 (1.4)&3.3 (-0.9)\\
&\texttt{FC DAgger 4} &1/5 (0/4)&2.0 (1.4)&5.6 (-1.4)\\
\bottomrule
\end{tabular}\\[11pt]
\end{table*}

\begin{figure*}
\begin{center}
\includegraphics[width=\textwidth]{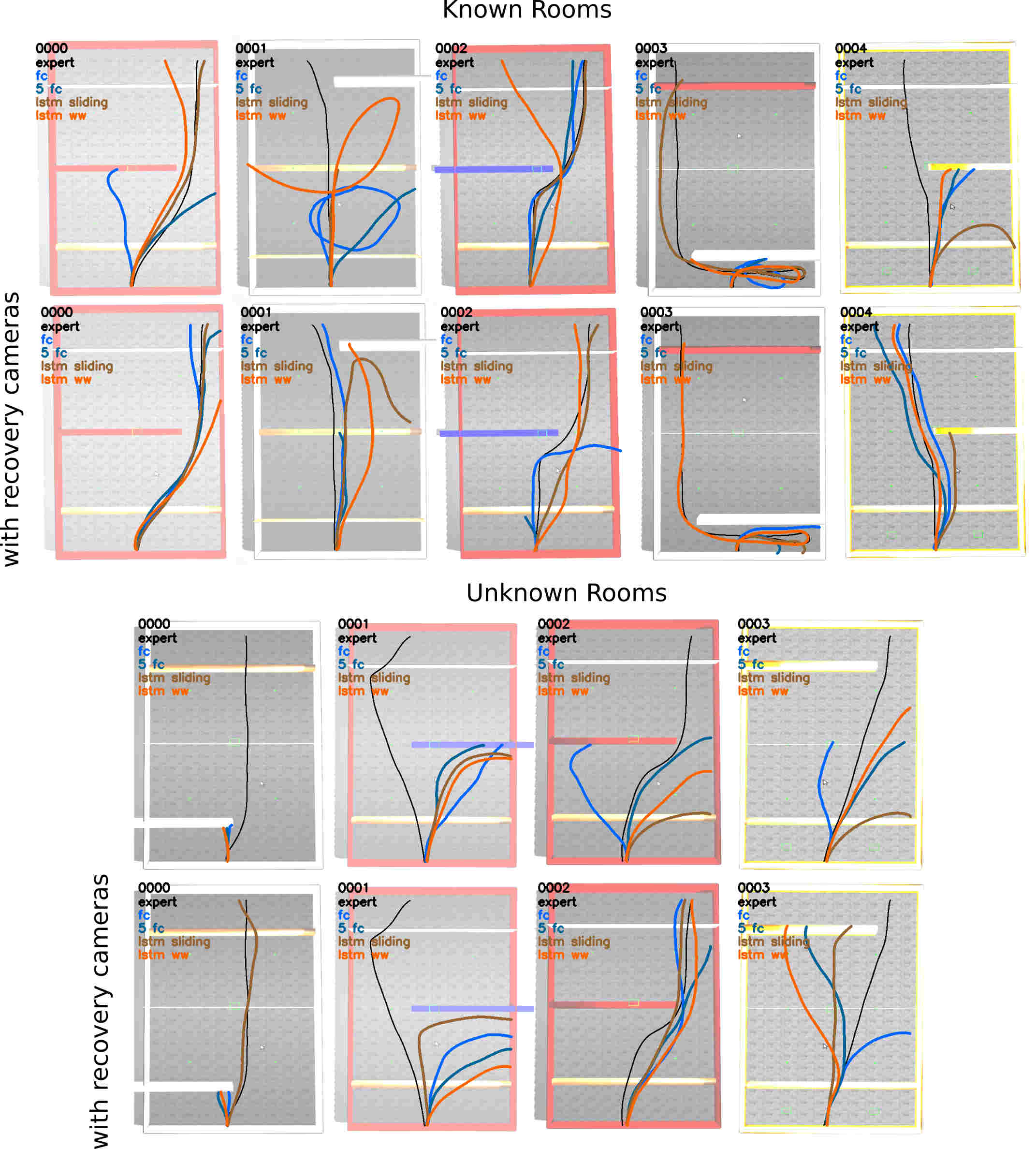}
\end{center}
\caption{The different trajectories flown through different rooms by networks with different architectures and training methods: light blue is FC, dark blue is 5-FC, brown is LSTM trained with S-BPTT, orange is LSTM trained with WW-BPTT and black is the expert. All the networks are trained without the recovery cameras in the 1st and 3th row and with recovery cameras in the 2nd and 4th row.}
\label{fig:arch-and-rec}
\end{figure*}

\subsection{FC vs 5-FC vs LSTM}
\label{sec:fcvslstm}
One of the main questions we want to address in this paper, is to see if memory helps for navigation control. We compare the different control architectures explained in section \ref{sec:controlnets}. The first half of table \ref{T2} and the first and third row in figure \ref{fig:arch-and-rec} show the results. 
As visible in table \ref{T2} the performance of the FC and the 5-FC is worse than the LSTM on all the different performance measures for the known rooms. This is not always the case for the unknown rooms (values between brackets). Though the imitation loss is always lower for the networks with memory, none of the networks succeed in crossing the room. The training data is not sufficient for any network architecture to perform reasonably well on the unknown rooms.

The first and third row of figure \ref{fig:arch-and-rec} show the trajectories for the five known and four unknown rooms. Also qualitatively it is visible for example in training room $0000$ and $0003$ that the trajectories of the LSTMs (brown and orange) are much closer to the experts trajectory (black) than the trajectories flown by the FC networks.

The main cost of training an LSTM instead of an FC control is the training time. The FC control is trained in less than 3 minutes while the LSTM can easily take more than 30 minutes trained with WW-TBPTT on the \textsl{Room Crossing Two} dataset on a 2G GPU. The online performance at test time happens at the same speed.

\underline{Conclusion:} There is a clear trend of LSTMs outperforming the FC control, which shows the usefulness of memory in navigation tasks. This is a very important result and a trend which is also visible in further experiments.


\subsection{Recovery data}
\label{sec:recovery}
It is important for the student not only to learn to copy the expert's behavior but also to learn how to recover from mistakes made. This is referred to as the state space shift. One way to deal with this, is by applying DAgger iterations as we do in subsection \ref{sec:dagger}. In this way the student makes a mistake and the expert annotates the proper control. This is a slow way of learning because the student can only learn one fatal mistake each test trajectory.

Another way of learning to recover is by providing this recovery data in the offline expert dataset.
There are two sources of potential drift for which recovery data can be provided.
The first source of drifting is a translation in the local z or y-axis perpendicular to the flying direction (local x-axis). It is usually not necessary to really recover from this drift, though the path should be adjusted during obstacle avoidance. In the training data of the Room Crossing One dataset the expert starts off from different global z and x positions. This results in a tube of trajectories as visible in figure \ref{fig:room}. If the student network drifts off the path it will still recognize the desired control from another trajectory of the expert closer to the current path.

The second source of drifting is in orientation. This is very plausible especially when the framerate is different during test time. The student control might turn a bit earlier or later than the expert. This results in a translation and an orientation difference.
In order to compensate for the orientation, we add a recovery camera on the left and the right of the center camera, as was also done by \cite{car}.
The RGB images obtained from the recovery cameras are annotated with controls that compensate for the different orientation. 
The compensation control steers the drone in more or less 2 seconds back in the original orientation.
During training the sequences from the right and the left camera are sampled in the same way as from the straight camera.

This introduces a recovery bias as two trajectories out of three are coming from a recovery camera which is looking in another direction than it is actually flying. This bias only manifests itself over several frames so not for the FC network which uses only 1 frame.

The performance of the networks on the \textsl{Room Crossing Two} dataset are listed in the table \ref{T2}. The performance of the FC increases much more than the performance of the LSTM. Resulting for instance in 3 successes for the FC control in the known rooms, while the network succeeded only 1 time without the recovery cameras. This will probably be because of the bias explained above.

With the recovery cameras, the amount of data is multiplied by a factor three. This has a similar impact on the training time. Besides handling the state space shift from expert to student, the data augmentation helps also against overfitting. This effect is visible in the sense that the FC and LSTM networks are capable of crossing some unknown rooms.

Given the positive effect on the performance, we use recovery cameras in the remaining experiments.

\underline{Conclusion:} Recovery cameras improve the performance significantly. The impact seems to be bigger for FC control networks than for LSTM networks probably due to the recovery bias.


\subsection{DAgger Iterations}
\label{sec:dagger}
\begin{figure*}
\begin{center}
\includegraphics[width=3cm]{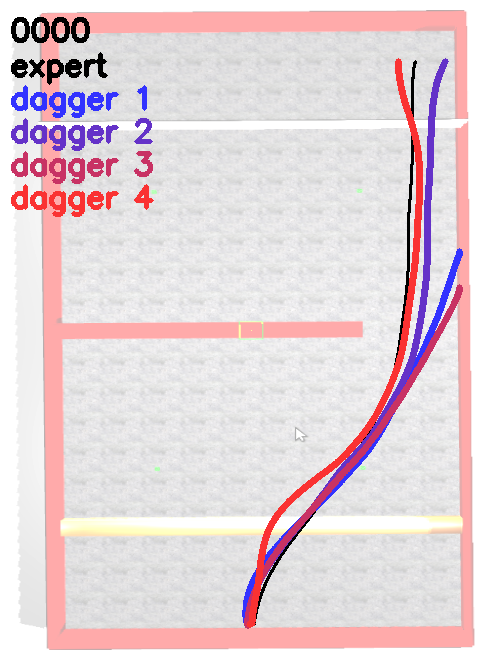}
\quad
\includegraphics[width=3cm]{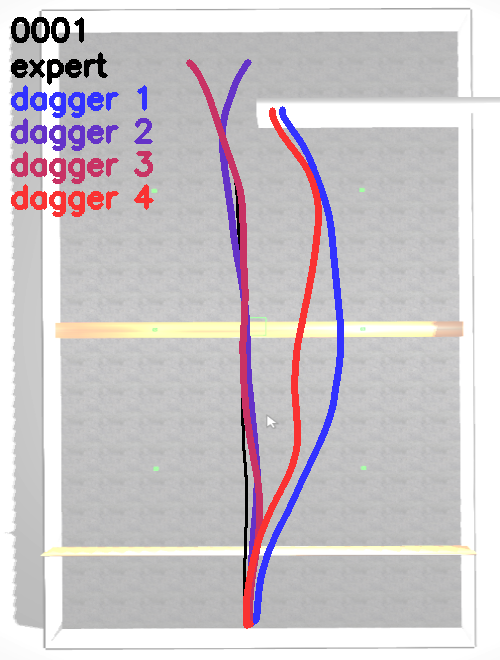}
\quad
\includegraphics[width=3cm]{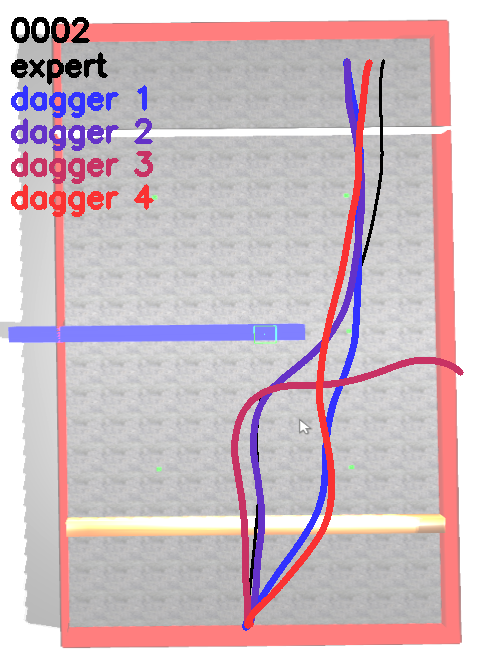}
\quad
\includegraphics[width=3cm]{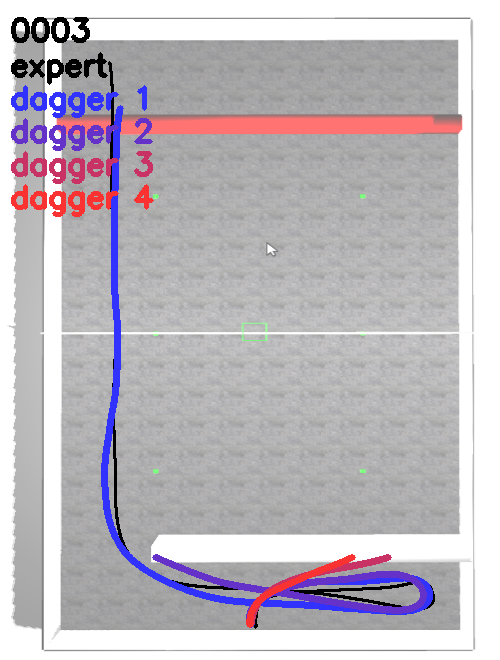}
\quad
\includegraphics[width=3cm]{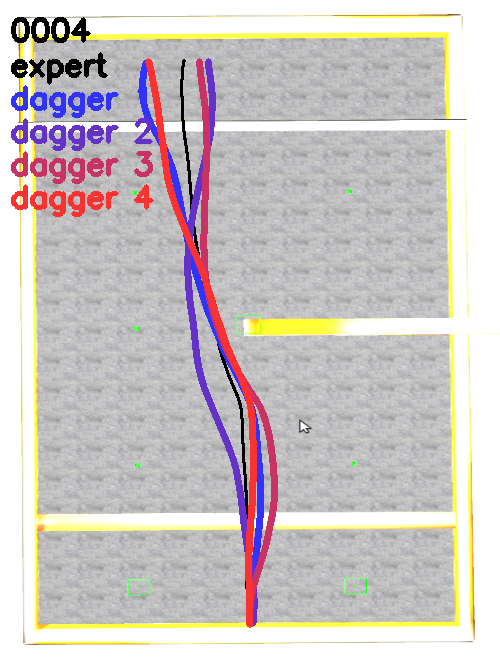}
\\
\includegraphics[width=3cm]{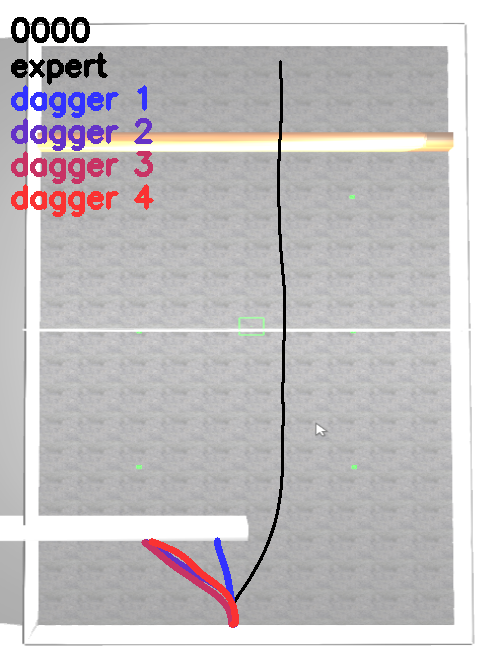}
\quad
\includegraphics[width=3cm]{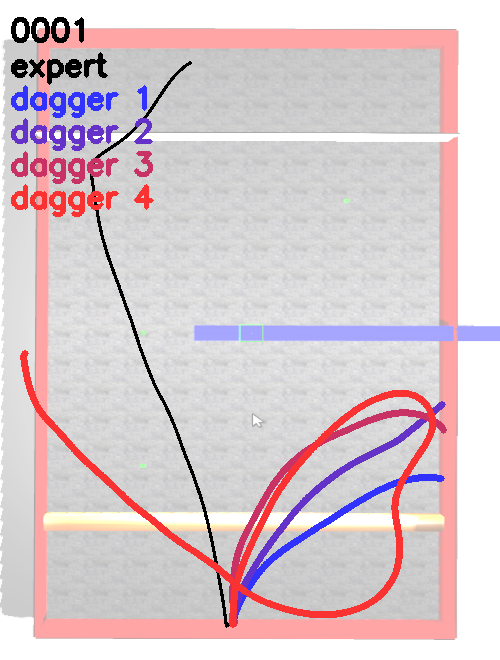}
\quad
\includegraphics[width=3cm]{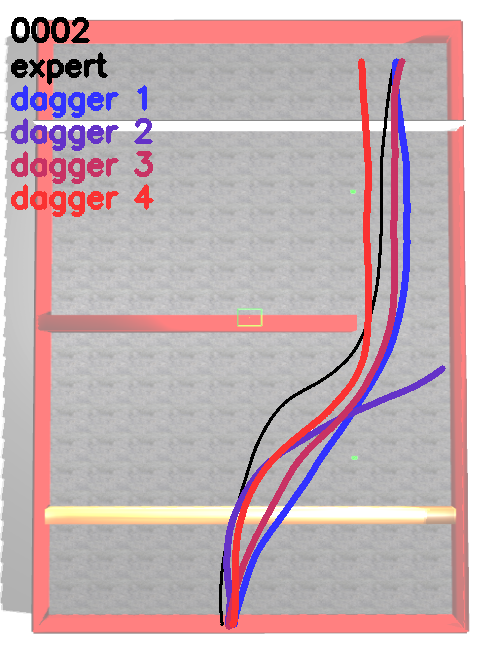}
\quad
\includegraphics[width=3cm]{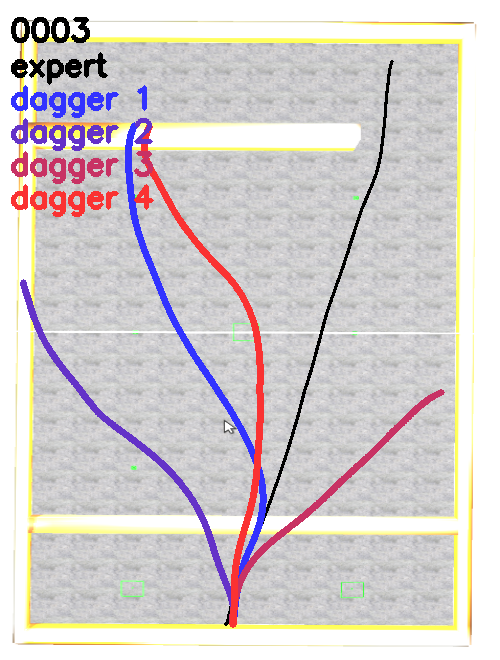}
\end{center}
\caption{Performance on known rooms controlled by the different networks iterated with DAgger.}
\label{fig:dagger-lstm}
\end{figure*}

Another way to compensate for the state space shift as explained before, is with the use of DAgger iterations. Figure \ref{fig:imitationmethod} shows the general setup in which the student iterates between online flying with an expert annotating and retraining on the aggregated dataset. 
Figure \ref{fig:dagger-lstm} shows for each known and unknown room of the \textsl{Room Crossing Two} dataset the different trajectories flown by an LSTM trained with WW-TBPTT over 4 different DAgger iterations from blue to red. In room $0000$ and $0001$ it is visible how the network can make a mistake at a later iteration even though it succeeded in passing this obstacle before. In room $0003$ the performance even seems to get worse at each DAgger iteration. Table \ref{T2} shows as well how the different performance measures of both the FC and the LSTM are not improving as expected. It is clear that applying DAgger iterations is not a waterproof way to deal with the state space shift in the context of deep control networks.


It seems that applying DAgger iterations introduces 2 biases which have a nefast influence on the performance. 

The first bias is due to the difference between the annotated control from the automated expert and the actual control applied by the network or student. This bias manifests itself only over different frames so it should not affect the FC control.

The second bias comes from the aggregated data in which the student follows different trajectories than the expert would like to, provoking annotated control from the expert that the expert would not apply in a normal situation. If the task is low level reactive obstacle avoidance, each frame with an obstacle in front is relevant for training the network. If the task is to navigate through a room, many different trajectories can be followed while the expert only prefers one. This results in confusing annotations steering the control networks in a wrong direction. 

The two biases can be limited by working with a larger training set made by the expert to keep the proportion of the biased training data made by the student low. 

Another way to increase the influence of the expert's dataset is by finetuning on a previously trained network. Instead of initializing the weights of the FC or LSTM network randomly at each DAgger iteration, one can initialize the weights of the network with the last network trained. In this way the network does not need to learn from scratch. The lowest 2 rows of table \ref{T2} shows how this improves the performance on the unknown rooms for the second DAgger iteration.

\underline{Conclusion:} DAgger iterations seem to be unreliable for dealing with the state space shift when applied on the \textsl{Room Crossing Two} dataset. This can be explained by 2 new biases, though further research is required. Fine-tuning the networks instead of training from scratch appears to have a slight positive influence.


\begin{figure}
\begin{center}
\includegraphics[width=0.5\textwidth]{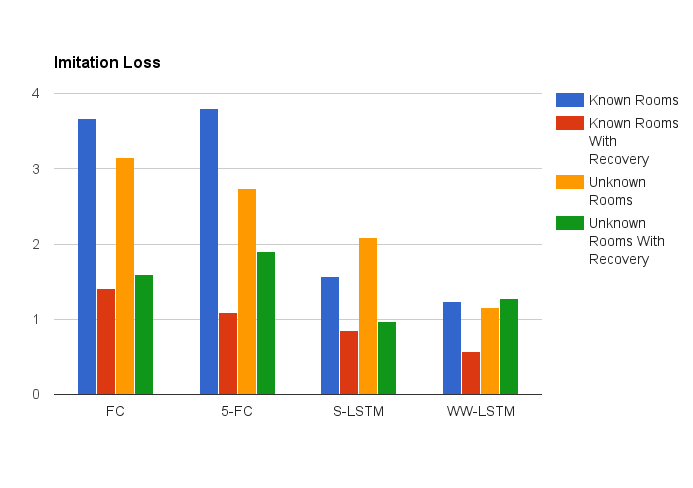}
\end{center}
\caption{The comparison of the average imitation loss on known and unknown rooms controlled by networks with different architectures and trained with different training methods, with and without recovery cameras. S-LSTM is trained with S-TBPTT and WW-LSTM is trained with WW-TBPTT.}
\label{fig:arch-and-rec-imitloss}
\end{figure}

\section{Discussion and Conclusion}
\label{sec:conclu}
After a series of experiments, numbers and figures, the most important preliminary conclusions are grouped in figure \ref{fig:arch-and-rec-imitloss} which shows the imitation loss on the \textsl{Room Crossing Two} dataset for our most important models. As the evaluation is done only in this basic simulated environment, we acknowledge limited reliability of the conclusions.

In this work we test how memory (in fig \ref{fig:arch-and-rec-imitloss}: LSTM two right vs FC two left groups) can help for deep neural networks in navigation control. In order to train an RNN, like an LSTM, it can be useful to decorrelate the training data with a method called window-wise truncated back propagation through time (WW-TBPTT). The method avoids the sequential bias of sliding truncated back propagation through time, though the higher variance and the calculations of the stored value makes the training process slower (in fig \ref{fig:arch-and-rec-imitloss}: right-most group).

We proposed a general imitation learning setup with an automated expert which uses extra sensor input. Here the setup is applied in a simulated environment though the setting is also applicable to a real environment with for example external motion capture systems as extra sensor input. The automated expert allows us to evaluate different trained networks with an imitation loss. The automated expert is able to perform a task a number of times for annotating recovery trajectories without the need of human interaction. It is implemented with behavior arbitration which makes it easy to implement an expert for different tasks. The expert can also be used to supervise the student automatically when running through different DAgger iterations. Recovery data from different trajectories and differently oriented cameras seemed to be crucial for the state space shift (in fig \ref{fig:arch-and-rec-imitloss}: red and green) while DAgger iterations seemed to be unreliable when applied to the \textsl{Room Crossing Two} dataset. Further research about the biases introduced by these methods is necessary.

Another important message from this work is the usefulness of pretrained networks. Only retraining the last FC layers of a convolutional network like Inception, trained on the Imagenet classification task, performs much better on the navigation control task than training the network end-to-end. End-to-end not only requires much more data, it also requires much more training time. This makes it often unfeasible to apply in real world situations in robotics.  

Finally this work gives general guidelines on how to apply imitation learning to deep neural networks for navigation control tasks. After publication, we will share the \textsl{Room Crossing One} and \textsl{Room Crossing Two} datasets which can be used as a benchmark for learning navigation control. The dataset serves as an indication of the required data for a task of a certain complexity. As a final good guideline we want to stress the need for different evaluation measures depending on the task and different visualizations in order to open the black box of deep learning.


\end{document}